\documentclass[letterpaper, 10 pt, journal, twoside]{IEEEtran}

\usepackage{amsmath,amssymb,amsfonts,url}
\usepackage{bm}
\usepackage{graphicx}
\usepackage{multirow}
\usepackage{booktabs}
\usepackage{color}

\usepackage{fancyhdr}
\usepackage{ifthen}
 
\ifCLASSINFOpdf
\else
\fi

\begin{document}
%
\title{Flying in Highly Dynamic Environments with End-to-end Learning Approach}

\author{Xiyu Fan$^{\dagger}$,
        Minghao Lu$^{\dagger}$,
        Bowen Xu,
        and Peng Lu
\thanks{Manuscript received: November 19, 2024; Revised January 7, 2025; Accepted February 24, 2025.}
\thanks{This paper was recommended for publication by Editor Jens Kober upon evaluation of the Associate Editor and Reviewers' comments.
This work was supported by General Research Fund under Grant 17204222, and in part by the Seed Fund for Collaborative Research and General Funding Scheme-HKU-TCL Joint Research Center for Artificial Intelligence.}
\thanks{$^\dagger$ Equal contribution.}
\thanks{The authotr are with the Adaptive Robotic Controls Lab (ArcLab), Department of Mechanical Engineering, The University of Hong Kong, Hong Kong, SAR, China {\tt\footnotesize fanxiyu, minghao0, link.bowenxu@connect.hku.hk; lupeng@hku.hk}}%
\thanks{Digital Object Identifier (DOI): 10.1109/LRA.2025.3547306} 
}

\hyphenation{op-tical net-works semi-conduc-tor}

\markboth{IEEE Robotics and Automation Letters. Preprint Version. Accepted February, 2025}
{Fan \MakeLowercase{\textit{et al.}}: Flying in Highly Dynamic Environments with End-to-end Learning Approach}
\maketitle
\begin{abstract}
Obstacle avoidance for unmanned aerial vehicles like quadrotors is a popular research topic. Most existing research focuses only on static environments, and obstacle avoidance in environments with multiple dynamic obstacles remains challenging. This paper proposes a novel deep-reinforcement learning-based approach for the quadrotors to navigate through highly dynamic environments. We propose a lidar data encoder to extract obstacle information from the massive point cloud data from the lidar. Multi frames of historical scans will be compressed into a 2-dimension obstacle map while maintaining the obstacle features required. An end-to-end deep neural network is trained to extract the kinematics of dynamic and static obstacles from the obstacle map, and it will generate acceleration commands to the quadrotor to control it to avoid these obstacles. Our approach contains perception and navigating functions in a single neural network, which can change from a navigating state into a hovering state without mode switching. We also present simulations and real-world experiments to show the effectiveness of our approach while navigating in highly dynamic cluttered environments. {Video: https://youtu.be/l4kLej8cUsQ.}
\end{abstract}



%
\IEEEpeerreviewmaketitle

\section{Introduction}

\IEEEPARstart{U}{nmanned} Aerial Vehicles (UAVs), particularly quadrotors, have ushered in a new era of possibilities across diverse applications in recent years, such as photography, logistics, and exploration \cite {racer, egoplanner, cinema}. Quadrotors have emerged as the predominant choice among UAVs owing to their adaptability and agility. Nevertheless, maneuvering quadrotors in cluttered environments typically demands the expertise of a skilled human pilot, leading to additional training requirements. Moreover, human response times can limit the full potential of quadrotors.


To address these challenges, autonomous obstacle avoidance techniques have been introduced. By integrating sensors like depth cameras or lidar, quadrotors can autonomously navigate through these environments. State-of-the-art obstacle avoidance approaches primarily rely on perception and path planning utilizing optimization algorithms \cite{egoplanner, mellinger2011minimum}, ensuring performance but often requiring substantial hardware resources for deployment. Furthermore, the latency of perception during high-speed flight can constrain performance.
\begin{figure}[t]
\centerline{\includegraphics[scale=0.12]{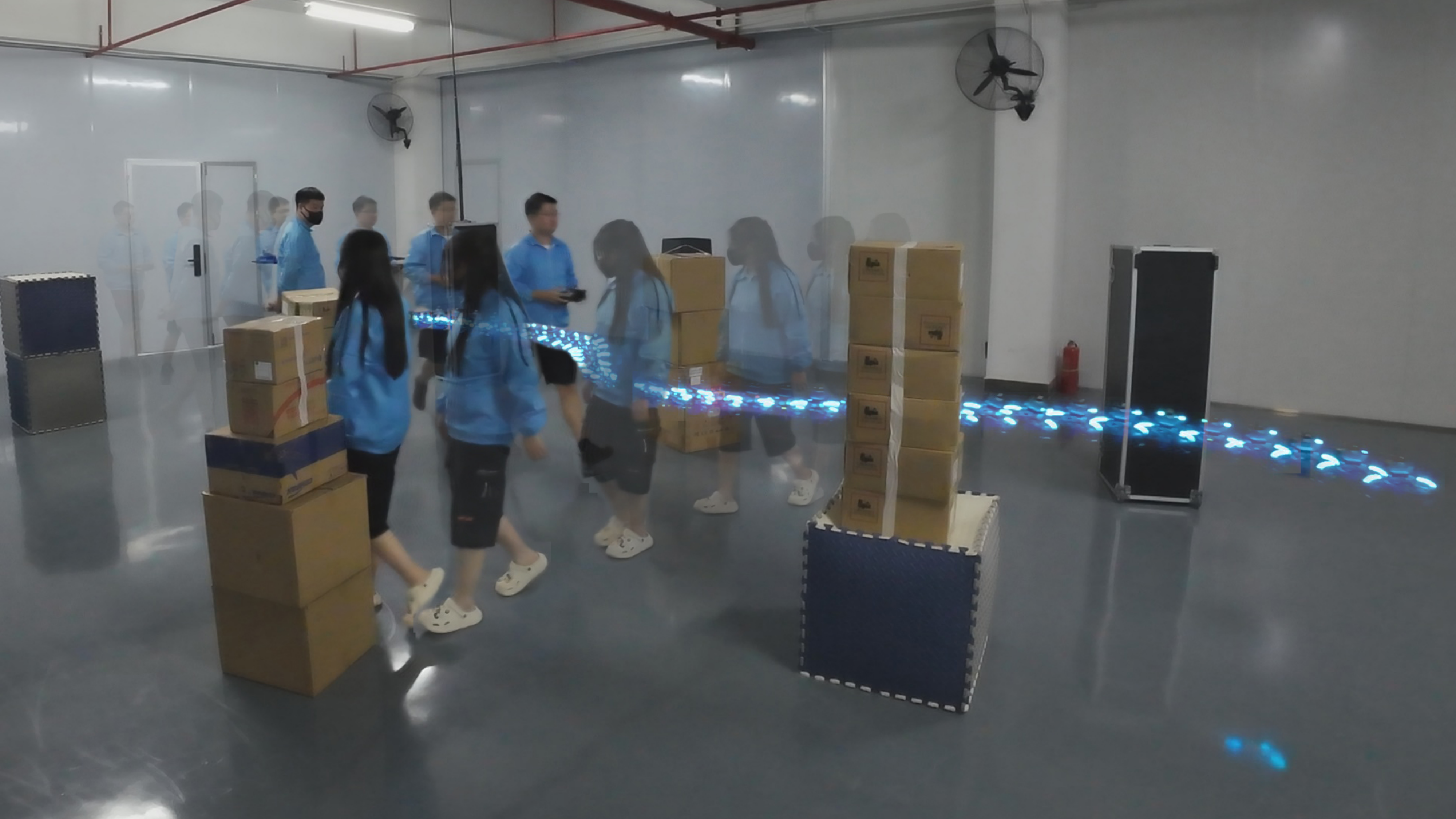}}
\caption{Our quadrotor flies through a dynamic cluttered environment, which contains both static obstacles and randomly walking pedestrians.}
\label{cover}
\end{figure}

In recent years, learning-based approaches have emerged as an alternative \cite{Song2023, loquercio2021learning, ross2013learning, kim2022towards}. Unlike optimization algorithms, neural networks are employed for obstacle avoidance, offering reduced latency and enabling faster responses compared to conventional optimization-based approaches.

However, the majority of current research only focuses on static environments. Obstacle avoidance in highly dynamic settings with fast-moving dynamic obstacles remains a challenge. Navigating through such environments not only poses planning complexities but also demands effective perception techniques.

In this paper, we introduce a novel end-to-end learning-based approach designed for obstacle avoidance in highly dynamic environments. Leveraging lidar for perception, the point cloud data captured by the lidar system is transformed into an obstacle map. Our approach employs Deep Reinforcement Learning (Deep-RL) to train a neural network specifically tailored for dynamic obstacle evasion.

The observation space of the neural network contains both the obstacle map and the current state of the quadrotor. By sending acceleration commands to the flight controller, the neural network directs the quadrotor to navigate and bypass both dynamic and static obstacles during flight.

Furthermore, we conduct a series of simulated and real-world experiments to compare our framework against the existing works. Our method showcases reduced response latency in comparison to optimization-based approaches, enabling rapid reactions to swiftly moving obstacles. In contrast to prevailing learning-based techniques, our methodology demonstrates superior performance in navigating through cluttered environments with a higher density of high-speed dynamic obstacles.

The key contributions of this research are outlined as follows:
\begin{itemize}
\item[1)] We introduce a novel lidar data encoding methodology. By compressing 3D point cloud data into a 2D obstacle map, our approach encapsulates the contours and dynamics of both static and dynamic obstacles.

{\item[2)] Our work proposes a deep reinforcement learning framework that incorporates both static and dynamic obstacles during the training phase. Our framework enables quadrotors to navigate through highly dynamic environments.}

\item[3)] We first realize the learning-based end-to-end dynamic obstacle avoidance. In contrast to rule-based approaches, our system exhibits obviously reduced latency on mobile computing platforms, enabling the quadrotor to evade high-speed obstacles moving at speeds of up to 6 $m/s$.
\end{itemize}

\section{Related Work}
\subsection{Static environments navigation}
Various approaches have been proposed to tackle the obstacle avoidance problem. In most of the works, real-time obstacle avoidance for mobile robots is realized by a two-piece framework comprised of mapping and planning. \cite{mellinger2011minimum} first realized optimization-based planning algorithm for trajectory generation of quadrotors. Subsequent research like \cite{egoplanner} proposed the integration of grid maps and optimization-based planning algorithms to facilitate real-time trajectory generation using onboard sensors like lidar or depth cameras. The methods enable the UAVs to fly in static cluttered environments robustly and avoid some slow-moving objects. However, the methods could not perform well in cluttered and highly dynamic environments, because their planning process did not include the observation and prediction of the moving objects.

Learning-based planners satisfy a faster response compared to conventional methods. \cite{Song2023} applied imitation learning to achieve high-speed flight in cluttered environments with depth cameras. In this work, a stated-based teacher policy was trained to fly in these environments with pre-calculated waypoints and various sensor data. The student policy shared the same action space with the teacher policy, but it only had the observation space containing depth image, velocity, and attitude data, and it would be trained by the teacher policy. This work realized high-agility flight in cluttered environments, but it is not compatible with highly dynamic environments. \cite{loquercio2021learning, ross2013learning, kim2022towards} also applied similar approaches.

\subsection{Dynamic obstacle avoidance}
In order to satisfy the requirements of navigating in environments with dynamic obstacles, researchers have developed some new methods.
\cite{fastsmall,activesense} applied image-based algorithm for dynamic obstacle perception, and added the kinematic of dynamic obstacles into the cost function of the optimization algorithm while planning. Such works perform well on fast-moving small objects, but due to the limit of image recognition algorithms, only specific objects with distinct characteristics can be tracked, which reduces their universality.
\cite{sr2020, mueggler2015towards} achieved fast-moving object avoidance based on the low latency of the event camera. However, event cameras are only sensitive to moving objects, making it difficult to fully perceive the information in complex environments.
To fill these gaps, \cite{lin2020robust, wang2021autonomous, FAPP} made further contributions to dynamic obstacle avoidance. In these works, dynamic obstacles would be segmented from the overall point cloud data by cluster algorithm, and a Kalman Filter was adopted to estimate the velocity of moving objects. The optimization algorithm would then generate the trajectory of the UAV according to the perception result. However, the point cloud-based dynamic perception methods were time-consuming, and the success rate would significantly decrease when dynamic objects move at a high speed. Besides, the classification of the dynamic point cloud was difficult to formulate by conventional methods, and some false detection conditions could not be eliminated. Moreover, solving the model predictive control (MPC) and polynomial optimization problems would be very time-consuming if the problem was very complex.

Furthermore, \cite{deeppanther} made contributions to learning-based dynamic obstacle avoidance. In this work, a conventional optimization-based planner served as the teacher policy, and imitation learning was applied to train a neural network as a planner, which brought a significant decrease in planning latency. However, it did not consider the perception of dynamic obstacles, and the proposed neural network was designed to avoid only a single dynamic obstacle.
\cite{drlvo} tailored a pedestrian avoidance methodology for ground-based robots, leveraging a combination of sensors such as lidar and RGB cameras for perception. This research utilized RGB camera and lidar inputs, enabling the system to monitor pedestrian movements based on a brief historical analysis of lidar data and the kinematic details derived from YOLOv3. While this approach proved effective for navigating congested environments by detecting and avoiding pedestrians, its scope is limited solely to pedestrian avoidance and does not extend to evading other categories of dynamic obstacles.
Analogously, \cite{wang2023curriculum} designed a learning-based approach for ground-based robot navigation, which also collects multiple historical scans for obstacle identification. In this work, semantic segmentation is applied to separate movable objects from the lidar point cloud. However, this work was designed for ground robots and relatively static environments, and it is not compatible with highly dynamic environments and quadrotors.
%
%
%
\begin{figure*}[t]
\centerline{\includegraphics[scale=0.3]{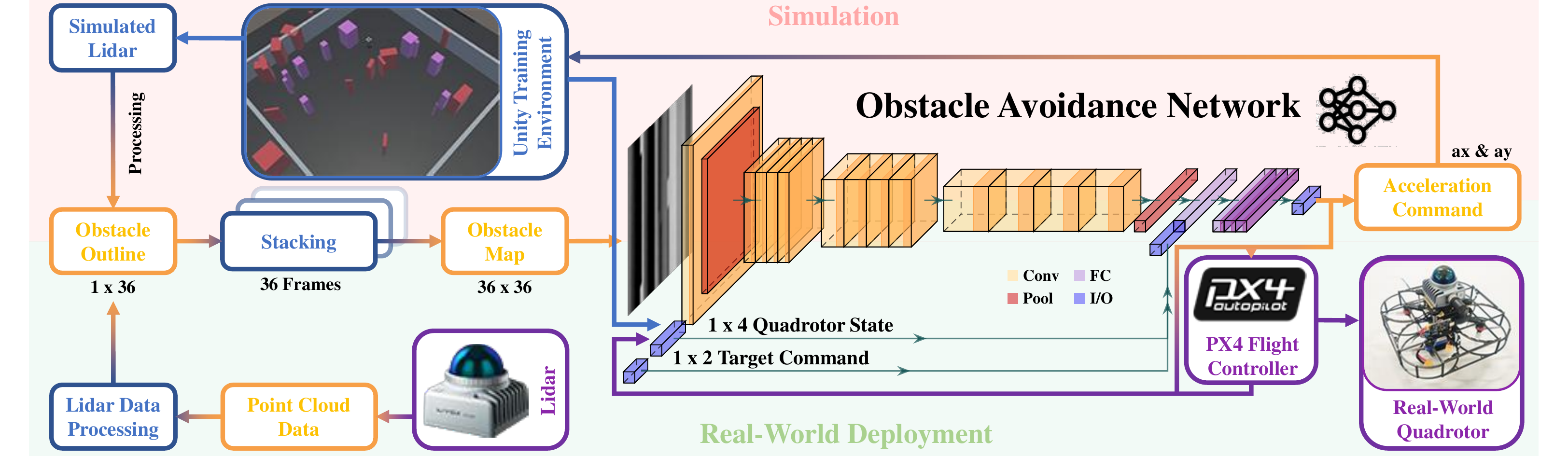}}
\caption{The overall architecture of our system. Raw data from lidar is processed into single-dimension range data, and stacked into a 36 × 36 obstacle map. The obstacle map is then fed into the encoder network, and the features extracted by the encoder are fed into the MLP together with the command and quadrotor state information. The output of the MLP is the acceleration command of 2 axes. 
}
\label{f:overallfig}
\end{figure*}

\section{Methodology}
In this article, we address the problem of planning the motion of a quadrotor to fly safely in highly dynamic environments with lidar sensing.

\subsection{Method Overview}
Our approach to dynamic obstacle avoidance while flying in cluttered environments contains a method for observing static and dynamic obstacle information from point cloud data, and a training strategy to train a neural network with deep reinforcement learning.

The overall architecture of our method is shown in Fig. \ref{f:overallfig}. The input of the system includes point cloud data of the environment, a set of vectors indicating the target position related to the quadrotor, and the state of the quadrotor including its velocity and acceleration command from the network in the last frame. The point cloud data will be encoded to compress obstacle information into a single-dimensional array to reduce data volume. We stack the compressed historical obstacle information in the previous 36 frames into a gray-scale image, and use an image encoder to extract the obstacle outlines and kinematic information of dynamic obstacles. Then, a Multi-Layer Perceptron (MLP) network is applied to generate horizontal accelerations according to the target command, quadrotor state observation, and obstacle information, therefore avoiding dynamic and static obstacles while navigating to the target position.


\subsection{Lidar Data Encoding}
This section introduces a methodology for encoding voluminous point cloud data from lidar into a 2-dimension array obstacle map, encompassing information on static and dynamic obstacles. This encoded data serves as a component of the input of the neural network.
\begin{figure}[t]
\centerline{\includegraphics[scale=0.1]{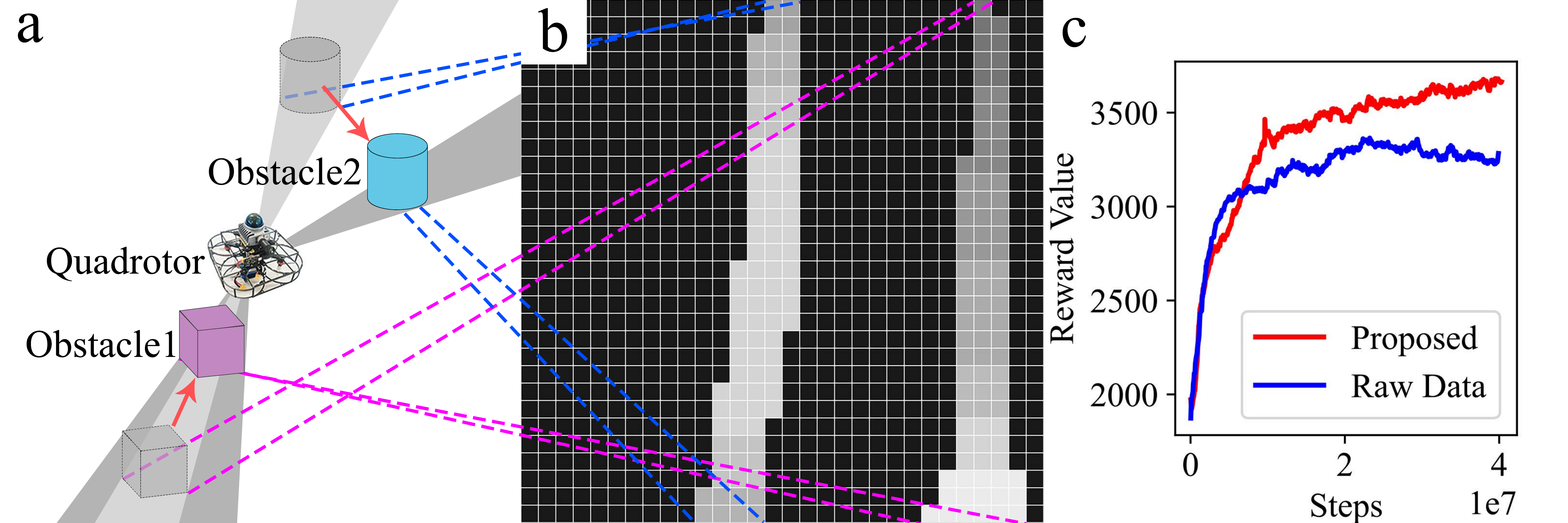}}
\caption{The sketch of obstacles and the corresponding obstacle map. (a) shows the kinematics of 2 moving objects, and (b) is the corresponding sketch of the obstacle map. The area in the obstacle map with higher gray levels represents an obstacle closer to the quadrotor. {(c) shows the learning curve of the agent using the proposed encoded lidar data and the raw lidar data.}}
\label{f:dynamicobs}
\end{figure}
The system perceives its surroundings through lidar, which furnishes a point cloud dataset comprising detected point positions. However, directly feeding this data into the neural network is impractical due to its varying size and substantial data volume. Let $_{B}P_{k}$ denote the point set obtained from lidar in the $k$th frame. To neutralize the attitude impact of the quadrotor, a $\mathbb{SO}(3)$ rotation transformation $_{B}^{W}T_{k}$ is applied to convert the point cloud data from the body frame $B$ to the horizontal global frame $W$, yielding the transformed point set $_{W}P_{k}$, and $_{B}^{W}T_{k}$ can be obtained from the onboard attitude estimator.

The scanning time of the lidar leads to significant delays when evading fast-moving dynamic obstacles. It usually takes a lidar 0.1 seconds to finish a complete scan, and incomplete scans might overlook smaller objects. Therefore, a sliding window is employed to process the point data. To be specific, a dynamic set of point clouds $\xi$ is defined, where $\xi \subseteq \{\underset{i \in [k-j,k]}{\bigcup} {_{W}P_{i}}\}$. This set retains point cloud data received from the lidar in the preceding $j$ frames, ensuring a stable observation of the point cloud while maintaining swift responses to rapidly moving small obstacles.

{For each point $p_i = (x_i, y_i, z_i) \in \xi$, its direction $\theta_i = \arctan\frac{x_i}{y_i}$ relative to the quadrotor can be determined. Here, we define the set of points in direction $s$ as $P_s = \{ p_i \in \xi \ | \ \|p_i\| < d_{max}, \ z_i \in [z_{min}, z_{max}], \ \theta_i \in [\frac{2\pi (s-1)}{n}, \frac{2\pi s}{n}] \}$, where $n$ is the angular resolution while generating the obstacle map, which is set to 36, and $d_{max}$ is the maximum distance that the obstacle map presents, which is set to 10 during training and real-world deployment.
To filter out the points that do not impede flight, like ground and ceiling, only points within the current flight level are kept, where $z_{min}=z_q-h$, and $z_{max}=z_q+h$. $z_q$ is the altitude of the quadrotor, and $h$ is the altitude threshold, which is set to 1 $m$ in real-world deployment.}
With the constant altitude of the quadrotor, the 3-dimension point cloud data can be compressed into a distance vector:
\begin{equation}
O(k)=[D_1(k),D_2(k),...,D_n(k)]^{\mathrm T}
\end{equation}
\begin{equation}{
D_s(k)= \frac{\mathop{\min}\{\|p_i\|, p_i \in P_s\}}{d_{max}}}
\end{equation}
{Furthermore, if $P_s = \varnothing$, we will let $D_s(k) = 1$. The distance vector $O(k)$ contains the distances of the nearest point to obstacles in each direction, which are denoted as $D_s(k)$. This format maintains a constant data size, occupies less storage, and provides adequate information for obstacle avoidance.}

{Inspired by \cite{drlvo}, we use a 2D array to express multi frames of lidar scan data.}
The historical distance vectors in the previous $m$ frames can be organized as:
\begin{equation}
M(k)=[O(k),O(k-1),...,O(k-m)]
\end{equation}
where $M(k)$ constitutes a 2D obstacle map comprising historical obstacle data from the above-mentioned frames, {and $m$ is set to 36 during training and real-world deployment}. Fig. \ref{f:dynamicobs} (b) illustrates the obstacle map derived directly from the vector $M(k),$ corresponding to the dynamic obstacles depicted in Fig. \ref{f:dynamicobs} (a). The two bands with lower gray levels represent the two dynamic obstacles respectively. Similarly, by analyzing these bands, the kinematics of all dynamic obstacles within the FOV can be observed.

{In order to validate the effectiveness of our lidar data encoding approach, a control group utilizing raw lidar data as input has been included while training the agent. Fig. \ref{f:dynamicobs} (c) illustrates the corresponding learning curve, which indicates that our proposed lidar data encoding approach leads to quicker convergence and achieves significantly higher final reward values compared to the control group.}

\subsection{Learning of dynamic obstacle avoidance}
In this subsection, we employ deep reinforcement learning to train a neural network capable of navigating the quadrotor through complex environments replete with dynamic obstacles.

\subsubsection{Problem Formulation}
The overall strategy involves training a universal policy leveraging deep reinforcement learning to control the acceleration of the quadrotor to fly through the obstacles. 

The problem is formulated as a Markov Decision Process (MDP), which is a discrete-time stochastic control process. The process can be formulated as a quadruple $M=(\mathcal{S},\mathcal{A},\mathcal{P},\mathcal{R})$, where $\mathcal{S}$ is the state space, $\mathcal{A}$ is the action space, $P(\mathcal{S},\mathcal{S}^\prime)$ is the probability that the action of the agent causes the state to transform from $S$ to $S^\prime$, and $\mathcal{R}(\mathcal{S},\mathcal{S}^\prime)$ is the reward that the agent received after the state to transform from $\mathcal{S}$ to $\mathcal{S}^\prime$. The reinforcement learning problem aims to adjust the policy $\pi$, in order to maximize the accumulated reward of the MDP $M$ with a discount ratio $\gamma$:
\begin{equation}
G_M(t)=\sum_{k=0}^\infty{(\gamma^kR_{t+k})}
\end{equation}

The state $S$ is the combination of the observation $\mathcal{O}$ and the environment parameters. The observation $\mathcal{O} \in [M(t),S(t),C(t)]$ contains the obstacle map $M(t) \in \mathbb{R}^{36\times36}$, the state of the quadrotor $S(t) \in \mathbb{R}^{1\times4}$, and the command input $C(t) \in \mathbb{R}^{1\times2}$. The obstacle map $M(t) \in \mathbb{R}^{36\times36}$ is constructed with multiple historical lidar scan results, processed through a Residual Network (ResNet) encoder \cite{resnet} to extract low-dimensional features. The state of the quadrotor $S(t) \in \mathbb{R}^4$ contains the current velocity and the acceleration output in the previous frame. Velocity data is acquired directly from the Unity environment during training and from the velocity estimator of the PX4 flight controller in real-world quadrotor deployment. The command input $C(t) \in \mathbb{R}^2$ comes from the user or the higher-level planner, representing the horizontal target position relative to the quadrotor. The action $\mathcal{A} \in \mathbb{R}^2$ comprises acceleration commands in the horizontal axes, executed by PX4 to adjust the attitude and thrust of the quadrotor during real-world deployment.

The reward $\mathcal{R}$ is computed through a series of reward functions based on the state of the quadrotor. Detailed information on the design of the reward functions will be stated in the subsequent subsection.

\subsubsection{Reward Function Design}
The reward function is designed to assess the actions of the quadrotor and provide the agent with feedback. Each component of the state contributes uniquely to the overall reward, influencing it in distinct ways.

The velocity reward, denoted as $r_v(t)$, serves to constrain the velocity of the quadrotor $v(t)$ within a specified range between the maximum velocity $v_{max}$ and the minimum velocity $v_{min}$. The computation of the velocity reward can be formulated as:
\begin{equation}
r_v(t)=r_{vmax}(t)+r_{vmin}(t)
\end{equation}
\begin{equation}
r_{vmax}(t)=\begin{cases}
e^{\|v(t)\|-v_{max}}-1& \|v(t)\|>v_{max}\\
0& \|v(t)\|\leqslant v_{max}
\end{cases}
\end{equation}
\begin{equation}
r_{vmin}(t)=\begin{cases}
e^{v_{min}-\|v(t)\|}-1& \|v(t)\|<v_{min}\\
0& \|v(t)\|\geqslant v_{min}
\end{cases}
\end{equation}

The progress reward, denoted as $r_p(t)$, prompts the quadrotor to navigate towards the designated goal position $p_{goal}$. The computation of the progress reward can be expressed as:
\begin{equation}
g(t)=\| p_{goal}-p(t)\|
\end{equation}
\begin{equation}
r_p(t)=e^{g(t)-g(t-1)}-1
\end{equation}
where $p(t)$ is the position of the quadrotor.

The jerk reward, denoted as $r_j(t)$, acts as a punishment against impractical maneuvers resulting from sudden changes in acceleration $a(t)$ of the simulated quadrotor. The computation of the jerk reward can be formulated as:
\begin{equation}
r_j(t)=e^{\|a(t)-a(t-1)\|}-1
\end{equation}

The obstacle avoidance reward, denoted as $r_o(t)$, plays a crucial role in penalizing the proximity of the quadrotor to static obstacles. Introducing a safety distance $d_s$, we utilize a distance function to assess the distance from the quadrotor to obstacles at a given position. The computation of the obstacle avoidance reward can be expressed as:
\begin{equation}
r_o(t)=\begin{cases}
e^{d_s-d(t)}-1& d(t)\leqslant d_s\\
0& d(t) > d_s\label{con:ESDF}
\end{cases}
\end{equation}
where $d(t)$ is the distance from the quadrotor to the closest point on the closest obstacle.
\begin{figure}[t]
\centerline{\includegraphics[scale=0.1]{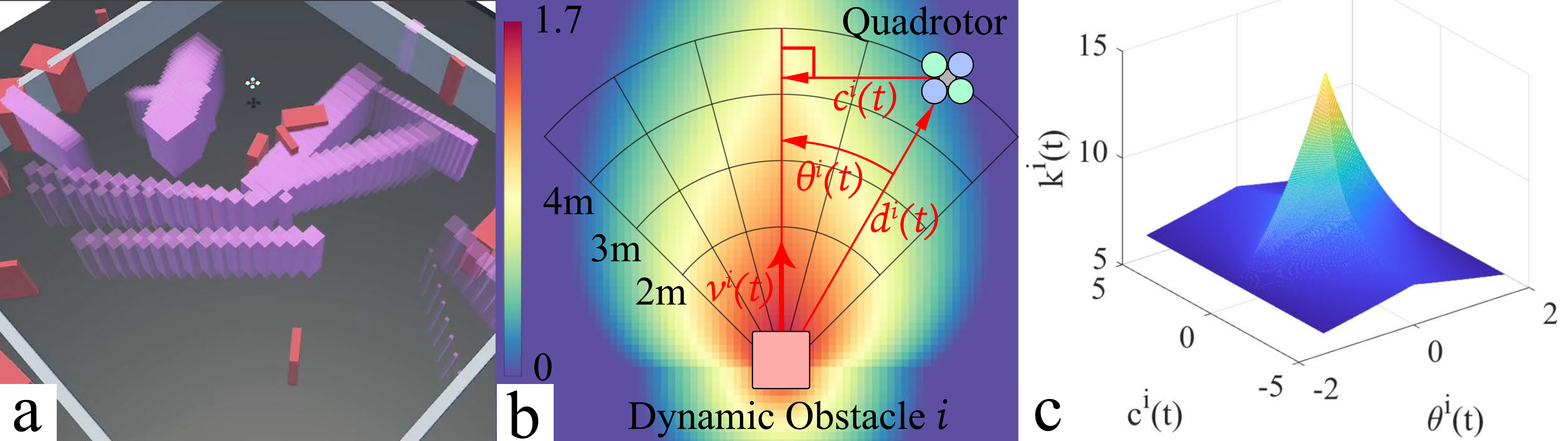}}
\caption{Our training environment in Unity. Static and dynamic obstacles are randomly generated. (a) is the training environment in Unity. (b) shows the dynamic obstacle reward $r^i_d(t)$ in Eq. \eqref{e:dyn_reward}. (c) demonstrates the dilation ratio $k^i(t)$ in Eq. \eqref{e:dilation}. The dynamic obstacle is moving at the speed of 5 $m/s$ in the calculation of the reward function in (b) and the dilation ratio in (c).}
\label{f:field}
\end{figure}

The dynamic obstacle reward, denoted as $r_d(t)$, aims to prompt the quadrotor to keep out from the future trajectory of dynamic obstacles. A dilation ratio $k^i(t)$ is employed to amplify the distance reward in the direction of movement of the dynamic obstacle. Assuming there are $n$ dynamic obstacles within the training environment, the dilation ratio for dynamic obstacle $i$ can be calculated with:
{
\begin{equation}
k^i(t)=\begin{cases}
1+\|v^i(t)\|(1-\frac{2\theta^i(t)}{\pi})e^\frac{1}{1+c^i(t)}& 0 \leqslant \theta^i(t)\leqslant \frac{\pi}{2}\\
1& \frac{\pi}{2}< \theta^i(t) \leqslant \pi
\end{cases}
\label{e:dilation}
\end{equation}}%
where $v^i(t)$ represents the linear velocity of dynamic obstacle $i$ under the global coordinate frame, $c^i(t)$ indicates the clearance distance from the quadrotor to the extended line of the velocity vector of this dynamic obstacle, and $\theta^i(t)$ denotes the included angle between the velocity vector of the obstacle and the vector from the obstacle to the quadrotor ({see Fig. \ref{f:field} (b)}). {Fig. \ref{f:field} (c) illustrates the curve of the dilation ratio when $v^i(t)=5$.}
The overall dynamic obstacle reward $r_d(t)$ can be computed as:
\begin{equation}
r_d^i(t)=\begin{cases}
e^{d_s-\frac{d^i(t)}{k^i(t)}}-1& \frac{d^i(t)}{k^i(t)}\leqslant d_s\\
0& \frac{d^i(t)}{k^i(t)} > d_s
\end{cases}
\label{e:dyn_reward}
\end{equation}
\begin{equation}
r_d(t)=\sum_{i=0}^n{r_d^i(t)}
\end{equation}
where $d^i(t)$ indicates the distance from the quadrotor to dynamic obstacle $i$. Fig. \ref{f:field} (b) illustrates an instance of dynamic obstacle reward.

The hovering reward, denoted as $r_h(t)$, gives additional rewards to the agent as the quadrotor approaches the goal point and its distance to the goal point is within a threshold distance $g_h$. It encourages the quadrotor to sustain a stable hover over the goal point until a new goal is designated. The computation of the hovering reward can be expressed as:
\begin{equation}
r_h(t)=\begin{cases}
e^{g_h-g(t)}-1& g(t)\leqslant g_h\\
0& g(t) > g_h
\end{cases}
\end{equation}

The total reward $\mathcal{R}$ at time $t$ is given by:
{
\begin{equation}
\begin{split}
\mathcal{R}=r_b - k_a \|a(t)\| - k_v r_v(t) - k_g g(t) - k_p r_p(t)  \\ - k_j r_j(t) - k_o (r_o(t) + r_d(t)) + k_h r_h(t)
\end{split}
\end{equation}}%
{where $k_a$, $k_v$, $k_g$, $k_p$, $k_j$, $k_o$, and $k_h$ represent the weights assigned to each component of the reward function, $a(t)$ is the acceleration of the quadrotor.} Additionally, $r_b$ denotes the fundamental reward value the agent receives in each step if the quadrotor has not collided with any obstacles.

\subsubsection{Policy Training}
{Unity is a powerful engine that facilitates kinematics simulation and provides relevant interfaces. With the Unity Machine Learning Agents Toolkit (ML-Agents), it is very convenient to build up a reinforcement learning environment by calling the related APIs, so we choose it as the training platform.}
The policy is trained with the Proximal Policy Optimization algorithm (PPO) \cite{ppo}, where the Actor network is defined as Fig. \ref{f:overallfig}, and the Critic network is embedded in ML-Agents. We employ 4 parallel agents to train the policy, and the training timescale has been set to 10.

In Fig. \ref{f:field} (a), the training environment is depicted, where each quadrotor operates within a square arena. At the beginning of every training episode, the side length of the arena is randomly set between 10 $m$ to 20 $m$. Simultaneously, static obstacles with randomized positions, rotations, and scales are generated. Additionally, a random goal position, maintaining a safe distance of 1 $m$ from any of the static obstacles, is specified. 5 dynamic obstacles, ranging in scale from 0.1 $m$ to 1 $m$, appear randomly at the edges of the arena. These dynamic obstacles move at random speeds from 1 $m/s$ to 6 $m/s$, and will be reset if collide with the edges of the arena. When a dynamic obstacle has been reset, it possesses a $50\%$ probability of moving towards the position of the quadrotor or moving in a random direction otherwise.

Each training episode has a step limit of 2000. If this limit is reached, the current episode concludes. In the event that a collision occurs before reaching the step limit, an additional punishment will be applied to the reward function, and the training episode will be terminated earlier.


\section{Experiments}
In this section, we showcase our simulation tests and real-world experiments aimed at validating the effectiveness of the method proposed in this paper.

\begin{figure*}[t]
\centerline{\includegraphics[scale=0.25]{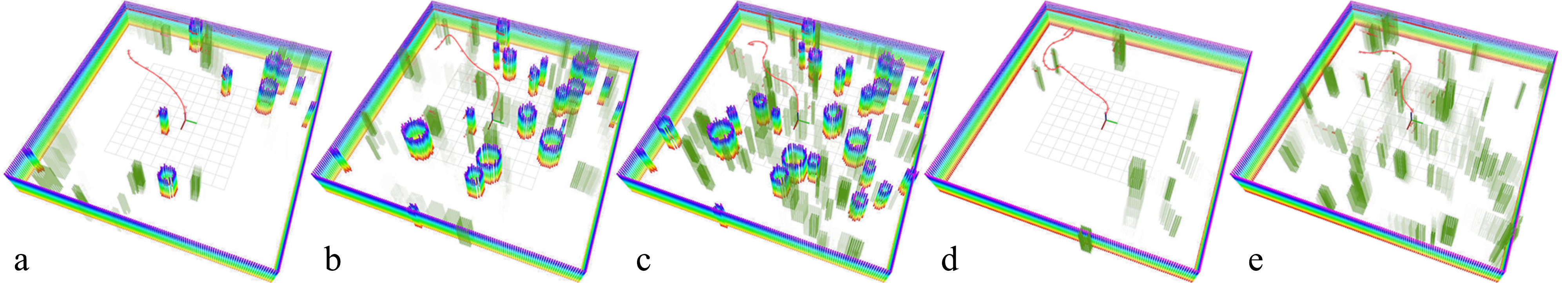}}
\caption{The quadrotor maneuvers through 5 distinct dynamic and cluttered environments in simulation. The scenarios labeled (a) to (e) correspond to scenarios 1 to 5 outlined in Table \ref{t:planning comp1}, respectively.}
\label{f:sim}
\end{figure*}

\subsection{Evaluations in Simulation}
To evaluate and compare the performance of our methodology against existing approaches, we design 5 distinct simulation environments for benchmarking purposes. Each environment is confined within a $20 m \times 20 m$ test ground enclosed by walls. Dynamic obstacles within these test grounds are characterized by random velocities ranging from 0 $m/s$ to 4 $m/s$, moving in random directions. In addition, the movement direction and velocity of dynamic obstacles underwent continual changes.

Illustrations of the simulation environments are shown in Fig. \ref{f:sim}. We execute the simulations on a desktop system featuring an i5-13600KF CPU and an RTX 4060 Ti GPU, operating on Ubuntu 20.04. For each scenario, we designate 10 random target points and assess the success rate of reaching these points without colliding with obstacles. Additionally, we record the average processing time, indicating the inference time for learning-based approaches or the combined perception and planning duration for optimization-based methods.

In our comparative analysis, we establish control groups, which include a traditional optimization-based method outlined in \cite{FAPP}, as well as learning-based methods detailed in \cite{drlvo} and \cite{uavdqn}.
\cite{drlvo} was initially developed for unmanned ground vehicles (UGVs) and features a distinct action space compared to quadrotors. Moreover, \cite{drlvo} incorporates both lidar and RGB image data as input, which may not seamlessly align with our quadrotor application scenario. On the other hand, \cite{uavdqn} leverages the global map of the environment as input, disregarding the perception aspect entirely, thereby rendering it incompatible with point cloud data input.

To ensure a fair and reasonable comparison, we introduce modifications to \cite{drlvo} and \cite{uavdqn}, aligning them with our simulation environment. These modifications contain alterations in the observation and action spaces, alongside adjustments to the reward functions to suit these modifications. The adapted reward functions, observation space, and action space utilized are detailed in Tables \ref{t:reward learning} and \ref{t:observation learning} respectively.

\begin{table}
\centering
\caption{Reward Functions for Learning-based Methods}
\begin{tabular}{ccc}
   \toprule
   Component Name & DRL-VO \cite{drlvo} & Double-DQN \cite{uavdqn}\\
   \hline
   \multirow{1}{*}{Acceleration} 
        & $a(t)$ & $\|v(t)-v(t-1)\|$ \\
    \hline
   \multirow{1}{*}{Jerk} 
        & $\|a(t)-a(t-1)\|$ & / \\
    \hline
   \multirow{2}{*}{Velocity} 
        & $e^{\|v(t)\|-v_{max}}-1 $& / \\
        & $\text{ if } \|v(t)\|>v_{max}$& / \\
    \hline
   \multirow{1}{*}{Goal Distance} 
        & $\|p(t)-p_{goal}\|$ & $\|p(t)-p_{goal}\|$ \\
    \hline
   \multirow{1}{*}{Goal Reaching} 
        & $e^{\|p(t)-p_{goal}\|-1}-1$ & $e^{\|p(t)-p_{goal}\|-1}-1$ \\
   \bottomrule
\end{tabular}
\label{t:reward learning}
\end{table}

\begin{table}
\centering
\caption{Observation and Action Space for Learning-Based Methods}
\begin{tabular}{ccc}
   \toprule
   Component Name & DRL-VO \cite{drlvo}& Double-DQN \cite{uavdqn}\\
   \hline
   \multirow{1}{*}{Target} 
        & 1*2 vector & 1*2 vector\\
    \hline
   \multirow{1}{*}{Velocity} 
        & 1*2 vector & 1*2 vector\\
    \hline
   \multirow{1}{*}{Acceleration} 
        & 1*2 vector & /\\
    \hline
   \multirow{2}{*}{Input} 
        & 1*36*36 & 1*36*36\\
        & stacked lidar data & stacked lidar data\\
    \hline
   \multirow{2}{*}{Action Space} 
        & 1*2 & 9*1\\
        & linear acceleration & 8 moving directions\\
   \bottomrule 
\end{tabular}
\label{t:observation learning}
\end{table}

\begin{table}[t]
\setlength{\tabcolsep}{5pt}
\centering
\caption{Dynamic Obstacle Avoidance Benchmark Comparison in Environments with Different Obstacle Densities}
{
\begin{tabular}{cccccc}
   \toprule
   \textbf{Scenario} & \textbf{Method} & \textbf{$\eta(\%)$} & \textbf{$t_p(ms)$} & \textbf{$v_a(m/s)$} & \textbf{$R_l$}\\
   \hline
   
 \multirow{3.5}{*}{10 dynamic obstacles}
 & Ours &\textbf{100} & \textbf{3.4} & \textbf{4.1} & 10.6\\
 \multirow{3.5}{*}{and 10 static obstacles} 
 & \cite{FAPP}& 70 & 11.8 & 1.9 & \textbf{8.5}\\
 & \cite{drlvo}& 40 & 5.4 & 1.4 & 11.8\\
 & \cite{uavdqn}& 50 & 5.2 & 2.0 & 25.4\\
   \hline
   
 \multirow{3.5}{*}{20 dynamic obstacles}
 & Ours & \textbf{70} & \textbf{3.8} & \textbf{4.5} & \textbf{9.4}\\
 \multirow{3.5}{*}{and 20 static obstacles}
 & \cite{FAPP}& 50 & 20.7 & 1.8 & 10.3\\     
 & \cite{drlvo}& 40 & 5.8 & 1.5 & 10.3\\
 & \cite{uavdqn}& 10 & 5.9 & 2.0 & 12.5\\
   \hline 
   
 \multirow{3.5}{*}{40 dynamic obstacles}
 & Ours & \textbf{50} & \textbf{5.2} & \textbf{3.8} & 13.0\\
 \multirow{3.5}{*}{and 30 static obstacles}
 & \cite{FAPP}& 30 & 35.0 & 1.5 & 9.4\\ 
 & \cite{drlvo}& 20 & 6.9 & 1.1 & \textbf{8.8}\\
 & \cite{uavdqn}& 0 & 7.1 & / & /\\
 \hline
 
 \multirow{4.5}{*}{10 dynamic obstacles}
 & Ours & \textbf{100} & \textbf{2.9} & \textbf{4.8} & \textbf{8.6}\\
 & \cite{FAPP}& \textbf{100} & 10.7 & 1.7 & 8.9\\
 & \cite{drlvo}& 90 & 5.0 & 1.4 & 13.7\\
 & \cite{uavdqn}& 90 & 4.7 & 2.0  & 16.7\\ 
 \hline 
 
 \multirow{4.5}{*}{40 dynamic obstacles}
 & Ours & \textbf{80} & \textbf{3.5} & \textbf{3.8} & 9.5\\
 &\cite{FAPP}& 40 & 25.6 & 1.8 & \textbf{9.3}\\ 
 & \cite{drlvo}& 60 & 5.5 & 1.7 & 10.7\\
 & \cite{uavdqn}& 20 & 5.4 & 2.0 & 9.7\\
 \bottomrule
\end{tabular}
}
\label{t:planning comp1}
\end{table}
\begin{table}[h]
\centering
\caption{Evaluation Of Reward Functions}
{
\begin{tabular}{ccc}
   \toprule
   \multirow{1}{*}{Average Dynamic}
    &Original Reward & Without dynamic\\
    \multirow{1}{*}{Obstacle Speed}
    &function&obstacle reward\\
    \hline
   \multirow{1}{*}{Low (1 $m/s$)} 
        & $60\%$ & $50\%$ \\
    \hline
   \multirow{1}{*}{Medium (2 $m/s$)} 
        & $50\%$ & $30\%$ \\
    \hline
   \multirow{1}{*}{High (3 $m/s$)} 
        & $40\%$& $20\%$ \\
   \bottomrule
\end{tabular}}
\label{t:dyn_reward}
\end{table}

\begin{figure*}[t]
\centerline{\includegraphics[scale=0.2]{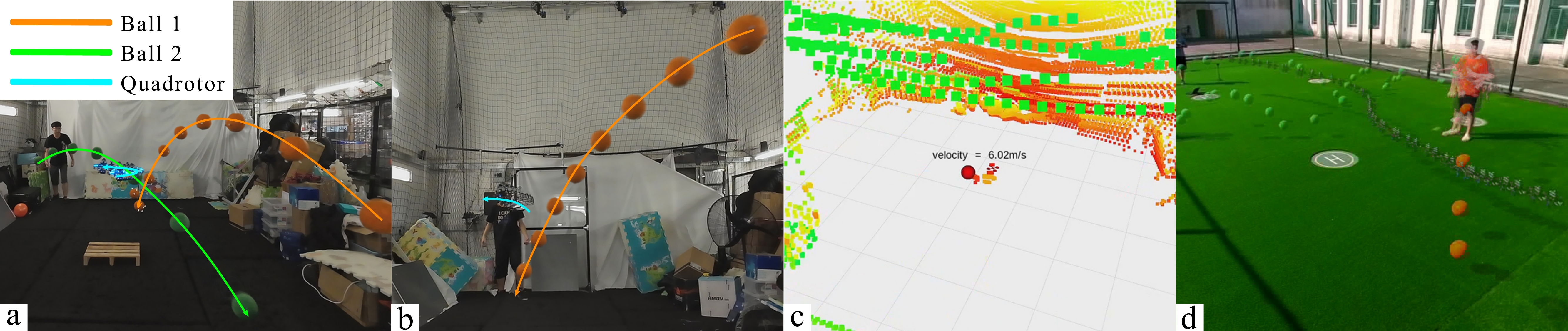}}
\caption{Here are some examples of our quadrotor successfully avoiding flying balls during real-world experiments. (a) presents the quadrotor avoiding multiple flying balls at the same time. In (b), the quadrotor successfully avoids a ball flying at a speed of 6 $m/s$, while (c) showcases the data captured by the lidar and the motion capture system during this evasion maneuver. (d) shows the quadrotor avoiding multiple balls while navigating.}
\label{f:ballspd}
\end{figure*}

\begin{figure*}[t]
\centerline{\includegraphics[scale=0.2]{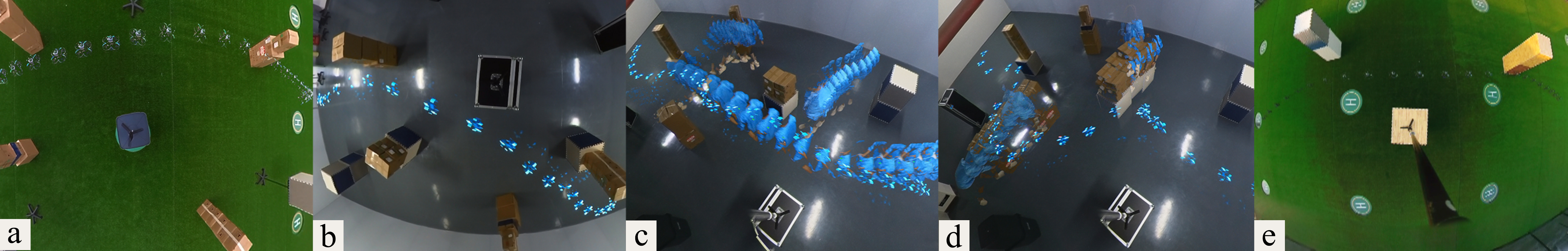}}
\caption{The trajectory of our quadrotor navigating through five different scenarios. (a) is set with static obstacles in low density, (b) is set with static obstacles in high density, (c) involves pedestrians walking around the obstacles, (d) contains pedestrians moving the obstacles, and (e) is set with a few static obstacles to test the high-speed performance.}
\label{f:alltrack}
\end{figure*}

Table \ref{t:planning comp1} demonstrates that our proposed approach exhibits superior efficiency and a higher success rate when navigating through environments populated with multiple high-speed dynamic obstacles. {$\eta$ presents the success rate of reaching the goal position without collision. $t_p$ denotes the processing time per step in learning-based approaches and presents the planning duration in optimization-driven approaches. $v_a$ is the average velocity during the flight. $R_l=10*l/\|P_g-P_s\|$ signifies the proportion between the path length and the direct path distance to the goal position, where $l$ denotes the path length, and $P_g$, $P_s$ represent the vectors of the initial and goal positions, respectively.}
While the above-mentioned methods achieve a decent success rate in environments with relatively low obstacle density, our approach maintains a decent success rate even as obstacle density increases, contrasting with the substantial success rate decline observed in the other methodologies. {Our approach additionally attains the maximum average velocity across all five scenarios and attains the shortest path length in two of these scenarios.}

The perception module in \cite{FAPP} experiences noticeable latency when processing point cloud data containing a large number of dynamic obstacles, leading to performance degradation. This latency significantly impacts the ability of the system to respond effectively in dynamic environments. Our processing time is shorter than both conventional and learning-based methods. Furthermore, the reward functions utilized in \cite{drlvo} and \cite{uavdqn} are not finely tuned for scenarios involving dynamic obstacles, resulting in a decline in performance in environments with high-speed dynamic obstacles. These reward functions hinder the ability of the agents to navigate efficiently in challenging dynamic environments.

{To assess the impact of the proposed reward function concerning dynamic obstacles, a series of ablation experiments are conducted. The reward function is modified by eliminating the dynamic obstacle reward $r_d(t)$ and applying the reward function for static obstacles $r_o(t)$ instead to compute the penalty attributed to dynamic obstacles. Three scenarios are established by varying the average speeds of the dynamic obstacles in the third scenario outlined in Table \ref{t:planning comp1}.}

{Table \ref{t:dyn_reward} illustrates the efficacy of the dynamic obstacle reward function in enhancing the success rate of navigation in highly dynamic environments. Furthermore, the enhancement in success rate attributed to the dynamic obstacle reward function becomes more pronounced as the average speed of dynamic obstacles increases.}
\subsection{Implementation of Real-world Experiments}
Our quadrotor platform design for real-world experiments features peripheral dimensions of 240 mm. The platform is equipped with propellers with all-around protection, ensuring the safety of experiments. This quadrotor platform is outfitted with a Morefine M6S onboard computer for neural network and odometry deployment, boasting an Intel N100 CPU and 12 GB of RAM. For sensing capabilities, it incorporates a Livox Mid-360 lidar with a wide field of view (FOV) of 360° horizontally and 59° vertically. The onboard computer runs on Ubuntu 20.04, and leverages Fast-Lio for odometry. In terms of flight control, the platform utilizes a mRo Pix-Racer Pro with PX4 firmware. With a total weight of 960 $g$, inclusive of a 1300 mAh battery, the platform offers a flight endurance time of 6.5 minutes. 
\subsection{High-speed Obstacle Avoidance}
Our methodology showcases a notable advantage in evading high-speed dynamic obstacles. To evaluate its efficacy in such scenarios, we conduct an experiment involving obstacles flying at a high speed. In this experiment, we assign the quadrotor a fixed target position and throw balls toward it while it is hovering. The quadrotor autonomously maneuvers to avoid these incoming balls, and the motion capture system records the positional data of both the quadrotor and the balls.

Throughout the experiment, our quadrotor is able to evade the ball flying at a speed up to 6 $m/s$, as illustrated in Fig. \ref{f:ballspd} (b) and (c). Utilizing a single neural network to handle various situations eliminates the need for the quadrotor to transition from a hovering state to a maneuvering state when faced with approaching obstacles. This reduces reaction times upon detecting dynamic obstacles within the field of view, enhancing performance in similar scenarios.

Our system is also able to process scenarios involving multiple high-speed dynamic obstacles. Fig. \ref{f:ballspd} (a) showcases the quadrotor deftly maneuvering to avoid two balls flying towards it simultaneously while it is hovering.

{
Furthermore, we also demonstrate that our quadrotor can avoid high-speed dynamic obstacles while navigating to a designated goal. Fig. \ref{f:ballspd} (d) shows that our quadrotor has successfully avoided multiple incoming balls while flying towards a goal at a speed of 2 $m/s$.
}

\subsection{Navigating in Cluttered Environments}
To evaluate the practical efficacy of our approach in cluttered dynamic environments, we design five distinctive real-world scenarios.

Initially, we establish a foundational scenario to validate the core navigation functionality. Diverse boxes of varying shapes and sizes were randomly positioned as obstacles on the test ground. The quadrotor is assigned a target position 10 meters ahead of its takeoff point. Fig. \ref{f:alltrack} (a) shows the trajectory of the quadrotor as it navigates through the obstacles without collision.

Subsequently, we heighten the challenge by introducing three additional obstacles to the test ground while maintaining the same target position. The quadrotor successfully maneuvers through these obstacles, but its velocity is reduced due to the increased obstacle density. The trajectory of this scenario is illustrated in Fig. \ref{f:alltrack} (b).

In the third scenario, we integrate three pedestrians into the environment, introducing dynamic obstacles among the static obstacles as the quadrotor flies through the test ground. 
Fig.\ref{f:alltrack} (c) showcases the trajectory of the quadrotor, and it is able to take evasive maneuvers when a pedestrian obstructs its path.

Expanding upon the complexity of the third scenario, pedestrians are tasked with altering the test ground layout by replacing obstacles in the flight path of the quadrotor. The quadrotor adeptly halts when facing an obstructing obstacle, subsequently recalibrating its route to avoid it. The trajectory of this experiment is depicted in Fig. \ref{f:alltrack} (d).

{
To highlight the efficacy of our methodology under high-speed settings, the fifth scenario involves sparse obstacle placement, challenging the quadrotor to swiftly navigate through them. The quadrotor achieves a notable speed of 6 $m/s$, surpassing the outcomes of 3 $m/s$ in \cite{zhou2022swarm}, and around 5 $m/s$ in \cite{schleich2022predictive}. The trajectory of the quadrotor in this scenario is depicted in Fig. \ref{f:alltrack} (e).
}

Across all of the scenarios, the quadrotor consistently maintains a hovering position at the designated target point upon arrival, awaiting commands for a new target destination. The series of scenarios demonstrate the adaptability and effectiveness of our navigation approach in maneuvering through dynamic cluttered environments.  

\section{Conclusion}
In this paper, we introduced a novel learning-based strategy designed for quadrotors to effectively navigate through highly dynamic and cluttered environments. Our methodology involved utilizing a lidar data processor to encode point cloud data into a range array. By stacking these range arrays into a 2D array, our neural network can discern dynamic obstacle features and execute appropriate evasive actions. We employed an end-to-end neural network for navigation, which controls the quadrotor by giving acceleration commands. A specialized reward function was applied to enhance its reaction against dynamic obstacles. 

Our approach showcased strong portability and excels across various dynamic environments, particularly in scenarios with high-speed obstacles. However, owing to the nature of end-to-end networks, the acceleration commands generated by the network may exhibit frequent fluctuations, and the current network framework was limited to obstacle avoidance within a horizontal plane. Future works may concentrate on expanding the maneuvering capabilities into three-dimensional space and enhancing overall stability.



%




\ifCLASSOPTIONcaptionsoff
  \newpage
\fi



%

%


\bibliographystyle{ieeetr}
\bibliography{ref}




\end{document}